\documentclass[letterpaper, 10 pt, conference]{ieeeconf}  % Comment this line out if you need a4paper

\IEEEoverridecommandlockouts                              % This command is only needed if 
\usepackage{graphicx}
\usepackage{booktabs}
\usepackage{subcaption}
\usepackage{adjustbox}
\usepackage{svg}
\usepackage{xcolor}
\usepackage{carshapes}
\usepackage{siunitx}
\usepackage{lipsum}
\usepackage{ifthen}
\usepackage{adjustbox}
\usepackage{float}
\usepackage{amsmath} % assumes amsmath package installed
\usepackage{amssymb}  % assumes amsmath package installed
\usepackage{dsfont}
\usepackage{tikz}
\usetikzlibrary{backgrounds}
\definecolor{cardinal}{HTML}{8C1515}
\definecolor{paloalto}{HTML}{175E54}
\definecolor{cardinalred}{RGB}{140,21,21}
\definecolor{coolgrey}{RGB}{83,86,90}
\definecolor{fog}{RGB}{218,215,203}
\definecolor{olive}{HTML}{4C580A}
\definecolor{lagunita}{HTML}{007C92}
\definecolor{darkbay}{HTML}{417865}
\definecolor{stone}{RGB}{127,119,118}
\definecolor{darkpoppy}{HTML}{D1660F}
\definecolor{plum}{HTML}{620059}
\definecolor{100black}{HTML}{2E2D29}
\definecolor{90black}{HTML}{43423E}
\definecolor{80black}{HTML}{585754}
\definecolor{70black}{HTML}{6D6C69}
\definecolor{60black}{HTML}{767674}
\definecolor{50black}{HTML}{979694}
\definecolor{40black}{HTML}{ABABA9}
\definecolor{30black}{HTML}{C0C0BF}
\definecolor{20black}{HTML}{D5D5D4}
\definecolor{10black}{HTML}{EAEAEA}
\definecolor{trafficyellow}{HTML}{F7B500}
\usetikzlibrary{positioning}
\usetikzlibrary{calc}
\usepackage[colorlinks = true,
            linkcolor = darkbay,
            urlcolor  = darkbay,
            citecolor = blue,
            anchorcolor = red]{hyperref}
\usepackage[capitalize]{cleveref}
\DeclareCaptionFormat{subfigureformat}{\adjustbox{valign=c}{#1#2#3}}

\newboolean{showcomments}
\setboolean{showcomments}{false} % Set to false to hide comments

\newcommand{\comment}[1]{%
    \ifthenelse{\boolean{showcomments}}%
    {\textcolor{red}{#1}}%
    {}%
}

\title{\LARGE \bf
SAVME: Efficient Safety Validation for Autonomous Systems Using Meta-Learning
}

\author{Marc R. Schlichting$^{1}$, Nina V. Boord$^{2}$, Anthony L. Corso$^{1}$, and Mykel J. Kochenderfer$^{1,2}$% <-this % stops a space
\thanks{$^{1}$Marc R. Schlichting, Anthony L. Corso, and Mykel J. Kochenderfer are with the Department of Aeronautics and Astronautics, Stanford University, Stanford, CA 94305, USA
        {\tt\small mschl@stanford.edu}}%
\thanks{$^{2}$Nina V. Boord and Mykel J. Kochenderfer are with the Department of Computer Science, Stanford University,
        Stanford, CA 94305, USA}%
}

\begin{document}

\maketitle
\thispagestyle{empty}
\pagestyle{empty}

%%%%%%%%%%%%%%%%%%%%%%%%%%%%%%%%%%%%%%%%%%%%%%%%%%%%%%%%%%%%%%%%%%%%%%%%%%%%%%%%
\begin{abstract}
Discovering potential failures of an autonomous system is important prior to deployment. Falsification-based methods are often used to assess the safety of such systems, but the cost of running many accurate simulation can be high. The validation can be accelerated by identifying critical failure scenarios for the system under test and by reducing the simulation runtime. We propose a Bayesian approach that integrates meta-learning strategies with a multi-armed bandit framework. Our method involves learning distributions over scenario parameters that are prone to triggering failures in the system under test, as well as a distribution over fidelity settings that enable fast and accurate simulations. In the spirit of meta-learning,  we also assess whether the learned fidelity settings distribution facilitates faster learning of the scenario parameter distributions for new scenarios.
We showcase our methodology using a cutting-edge 3D driving simulator, incorporating 16 fidelity settings for an autonomous vehicle stack that includes camera and lidar sensors. We evaluate various scenarios based on an autonomous vehicle pre-crash typology. As a result, our approach achieves a significant speedup, up to 18 times faster compared to traditional methods that solely rely on a high-fidelity simulator.

% Discovering potential failures in autonomous systems is a challenging yet vital part of building trustworthy AI. Falsification-based methods are commonly employed to assess the safety of such systems, but they often necessitate large numbers of highly accurate simulations. An acceleration of validation time can be achieved through two methods: identifying critical failure scenarios for the system under test and reducing simulation runtime. We propose a Bayesian approach that integrates meta-learning strategies with a multi-armed bandit framework. Our method involves learning distributions over scenario parameters that are prone to triggering failures in the system under test, as well as a distribution over fidelity settings that enable fast and accurate simulations. In the spirit of meta-learning,  we also assess whether the learned fidelity settings distribution facilitates faster learning of the scenario parameter distributions for new scenarios.
% We showcase our methodology using a cutting-edge 3D driving simulator, incorporating 16 fidelity settings for an autonomous vehicle stack that includes camera and lidar sensors. We evaluate various scenarios based on an autonomous vehicle pre-crash typology. As a result, our approach achieves a significant speedup, up to 18 times faster compared to traditional methods that solely rely on a high-fidelity simulator.

\end{abstract}

%%%%%%%%%%%%%%%%%%%%%%%%%%%%%%%%%%%%%%%%%%%%%%%%%%%%%%%%%%%%%%%%%%%%%%%%%%%%%%%%
%Introduction
\section{\textsc{Introduction}}
% Establishing trust in autonomous systems, such as self-driving vehicles, is a major challenge in current research. When autonomous systems fail at seemingly trivial tasks, their perceived safety is comparatively lower than that of human-operated counterparts performing the same tasks. \cite{nair2021sharing}.
One way of demonstrating the reliability of an autonomous system is through rigorous real-world testing, a process that requires substantial resources and is often economically infeasible \cite{kalra2016driving}. For this reason, simulations have become the method of choice in both research and industry for the validation of autonomous systems. Compared to real-world testing, simulations are faster, cheaper, and safer \cite{sovani2017simulation}.\begin{figure*}[!tb]
\centering
  \resizebox{\textwidth}{!}{\input{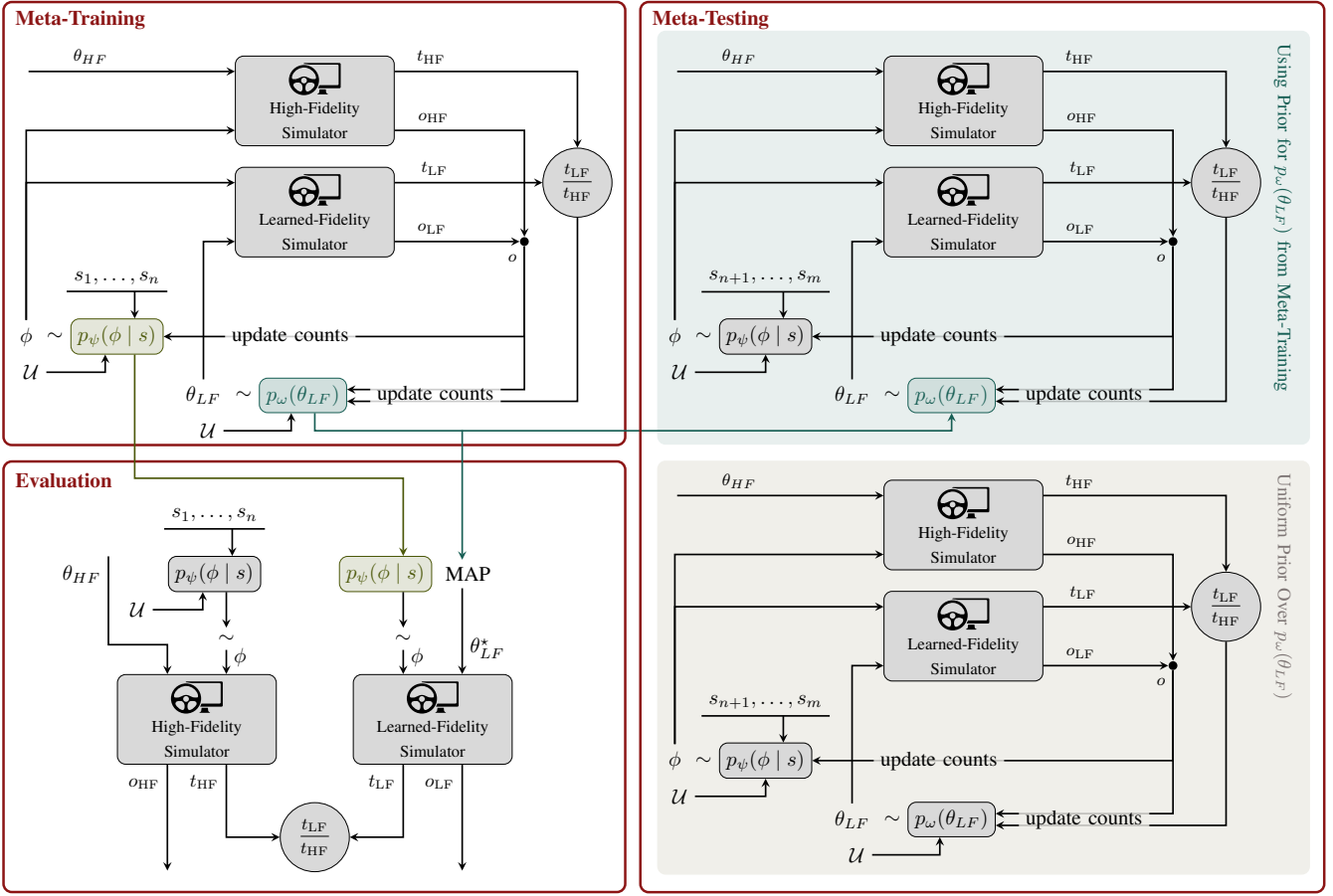}}
  \caption{Meta-learning framework for efficient safety validation.}
  \label{fig:extended_meta_learning}
\end{figure*}
\par  While empirical evidence supports the advantages of simulations over real-world testing, one fundamental question remains: \textit{How much can we trust simulations}? Safety validation relies on a vast number of simulations to find edge cases through falsification \cite{corso2020survey}. This leads to a dilemma: should the accuracy of the simulation or its runtime be prioritized? Many modern-day simulation tools give the user control over numerous settings that affect behavior, output, and compute cost. Such settings range from different equations of motion to numerical solvers and sensor models \cite{rosique2019systematic}. Depending on the system under test, different fidelity settings are more important than others in terms of arriving at accurate safety assessments. 
\par There are two important components of accelerating the validation process: speeding up the runtime and efficiently finding scenarios where the system under test fails. Many different approaches for finding failure scenarios have been studied in the literature. One such approach is black-box optimization \cite{kochenderfer2019algorithms}, which has been applied to safety validation of autonomous systems \cite{donze2010breach, annapureddy2010ant}. Two other similar approaches are path planning \cite{lewis2012optimal} and reinforcement learning \cite{koren2018adaptive}. A detailed review of methods is provided by Corso et al. \cite{corso2020survey}. These approaches require a simulation that is assumed to model reality.
% \par There are two important components of accelerating the validation process: learning fidelity settings that speed up runtime and learning scenario configurations where the system under test is more likely to fail. This means optimizing for specific scenario configurations of existing pre-crash typologies \cite{najm2007pre} rather than random sampling of scenario parameters. Many different approaches for finding failure scenarios have been studied in the literature. One such approach is black-box optimization \cite{kochenderfer2019algorithms}, which has been applied to safety validation of autonomous systems \cite{donze2010breach, annapureddy2010ant}. Two other similar approaches are path planning \cite{lewis2012optimal} and reinforcement learning \cite{koren2018adaptive}. A detailed review of methods is provided by Corso et al. \cite{corso2020survey}. These approaches require a simulation that is assumed to model reality.
\par Photorealistic simulators such as Carla \cite{dosovitskiy2017carla}, Airsim \cite{shah2018airsim}, or the products by Applied Intuition have been effective tools for developing and testing autonomous driving stacks. This realism, however, comes at the cost of increased computational resources required to run safety analyses. Since most simulators expose a number of fidelity settings to the user, it is possible to adapt those settings for specific needs. For example, an autonomous system stack without cameras does not need rendering. Unfortunately, this tuning process is non-trivial and requires domain expertise. For this reason, there are frameworks that make use of different fidelity settings by combining compute-intensive results from a high-fidelity simulator with computationally cheaper results from a low-fidelity simulator \cite{xu2014efficient}. This approach has been developed further to support more than two levels of fidelity and has also been applied to safety validation of autonomous systems \cite{beard2022safety}. Multi-fidelity approaches have been shown to perform well while keeping computational requirements low, but are limited to a finite number of fidelity setting combinations \cite{pellegrini2022multi}. In reality, however, simulators often offer dozens of parameters that determine the fidelity of a simulation. Consequently, it is infeasible to consider all possible combinations of settings.
\par Rather than combining the results from a limited number of simulators with expert-selected fidelity settings, our approach simultaneously learns the optimal fidelity settings (for an arbitrary simulator with many fidelity settings) while concurrently learning scenario configurations where the system under test is more likely to fail. The overall goal is to maximize the number of failures found in a given time. Our approach works with a combination of mixed continuous and discrete fidelity and scenario settings while taking the uncertainty of the outcome into account. To achieve this, we use a meta-learning framework where the task is learning the scenario parameters that lead to system failure. The optimized fidelity settings are considered as a prior that facilitate finding failures for new scenarios that have yet to be encountered during the training faster. The framework requires a high-fidelity simulator---providing the ground truth---and a learned-fidelity simulator during the training phase. Uncertainty is taken into account by framing the optimization problem as a multi-armed bandit problem using Bayesian model estimation and Thompson sampling.
\par We validate the feasibility of our approach using a cutting-edge 3D driving simulator. A total of 16 fidelity settings can be controlled, leveraging an autonomous vehicle stack comprising both camera and lidar sensors. The scenarios used for the experiments are derived from an autonomous vehicle-specific pre-crash typology \cite{liu2021crash}. Through our experiments, we demonstrate the capability of our framework to not only learn the probability distribution of failure likelihood across scenario parameters, but also the distribution for accurate and fast simulation results across the fidelity settings. Despite the computational overhead originating from the parallel use of the high-fidelity and learned fidelity simulator, we reach the break-even point---the time at which we found more failures compared to only using a high-fidelity simulator---before the end of the training phase. Through the parallel training, we reduce the average time to find a failure by a factor of 18. Furthermore, we demonstrate that the acquired distribution over the fidelity settings can be used as a warm start for learning the parameter distributions for a novel set of scenarios. Using this head start, we can accelerate the learning process up to two times over the parallel learning with uniform prior as described above.
\par This paper makes two key contributions. First, it presents a simulator-agnostic framework that enables us to find distributions over both scenario parameters that yield a high probability of failure and fidelity settings that yield a high probability of a fast runtime while maintaining accuracy. Second, these scenario-agnostic fidelity settings facilitate the accelerated learning of the distribution over scenario parameters for new scenarios, increasing the efficiency of validating new scenarios after training.  

\section{\textsc{Methodology}}
Our framework is based on meta-learning and Bayesian approaches for multi-armed bandits. This section explains how we combine both techniques to maximize the number of failures we find with constrained compute time. Let $s$ be an abstract scenario description and $\phi$ represent the scenario configuration, which is a vector that contains all values to create a runable instance of the scenario. The learned-fidelity settings---denoted by $\theta_{\mathrm{LF}}$---are the values of all the fidelity settings for the learned-fidelity simulator. The SAVME framework learns the scenario-specific distributions $p_\psi(\phi\mid s)$---with $\psi$ as parameters---at the same time as the distribution over the scenario-agnostic fidelity settings $p_\omega(\theta_{\mathrm{LF}})$, parameterized by $\omega$. A core component of our framework is the concurrent use of a high-fidelity simulator with fidelity settings $\theta_{\mathrm{HF}}$ and a learned-fidelity simulator with fidelity settings $\theta_{\mathrm{LF}}$. To be precise, we want to solve two constrained optimization problems:
\begin{equation}
\begin{aligned}
& \underset{\psi}{\text{maximize}}
& & \mathbb{E}_{\substack{s\sim p(s)\\ \phi\sim p_\psi(\phi\mid s)}}\left[\mathds{1}\left[\mathrm{sim}(s,\phi,\theta_{\mathrm{LF}}^\star,\theta_{\mathrm{HF}}^{\phantom{\star}})\in \mathcal{S}_\mathrm{scenario}\right]\right]\\
& \text{subject to}
& & \theta_{\mathrm{LF}}^\star=\underset{\theta_{\mathrm{LF}}}{\arg\max}~p_\omega(\theta_{\mathrm{LF}}),
\end{aligned}
\end{equation}
\vskip -0.3cm
\begin{equation}
\begin{aligned}
& \underset{\omega}{\text{maximize}}
 & & \mathbb{E}_{\substack{s\sim p(s)\\ \phi\sim p_\psi(\phi\mid s)}}\left[\mathds{1}\left[\mathrm{sim}(s,\phi,\theta_{\mathrm{LF}}^\star,\theta_{\mathrm{HF}}^{\phantom{\star}})\in \mathcal{S}_\mathrm{fidelity}\right]\right]\\
& \text{subject to}
& & \theta_{\mathrm{LF}}^\star=\underset{\theta_{\mathrm{LF}}}{\arg\max}~p_\omega(\theta_{\mathrm{LF}}),
\end{aligned}
\end{equation}
where \textsc{sim} is the function that simulates scenario $s$ with parameters $\phi$, using the high and learned-fidelity simulator with fidelity settings $\theta_{\mathrm{HF}}$ and $\theta_{\mathrm{LF}}^\star$, respectively. \textsc{sim} returns the outcome $o\in\{\mathrm{TP},\mathrm{TN},\mathrm{FP},\mathrm{FN}\}$ and the runtimes of the high and learned-fidelity simulators, $t_\mathrm{HF}$ and $t_\mathrm{LF}$, respectively:
\begin{equation}
    o,t_\mathrm{LF},t_\mathrm{HF}=\mathrm{sim}(s,\phi,\theta_\mathrm{LF}^\star,\theta_\mathrm{HF}^{\phantom{\star}}).
\end{equation}
For learning the scenario parameters, a success $\mathcal{S}_\mathrm{scenario}$ is defined as a failure of the system under test that is found using the high-fidelity simulator while for learning the fidelity settings, a success $\mathcal{S}_\mathrm{fidelity}$ is an outcome agreement between the high and learned-fidelity simulator in conjunction with a fast runtime. A more detailed description is given in \cref{sec:fidelity_learning,sec:scenario_learning} and summarized in \cref{tab:failure_success}. Throughout this work, we assume that $p(s)$ is uniform, meaning that all scenarios are equally likely. However, $p(s)$ can be an arbitrary distribution.
\subsection{Meta-Learning Framework}\label{sec:extended_meta_learning}
We partition the meta-learning framework into three phases: \textit{meta-training}, \textit{evaluation}, and \textit{meta-testing}. \Cref{fig:extended_meta_learning} shows these phases of the framework. 
\subsubsection{Meta-Training} 
For each training iteration, the scenario parameters $\phi$ for one of the randomly selected training scenarios $s_1,\ldots,s_n$ are sampled from $p_\psi(\phi\mid s)$. In a similar fashion, we sample the learned-fidelity settings $\theta_{\mathrm{LF}}\sim p_\omega(\theta_{\mathrm{LF}})$ as well. The high-fidelity settings $\theta_{\mathrm{HF}}$ are determined \textit{a priori} and are never changed throughout the training, evaluation, or testing process. During the meta-training phase, we run the same scenario using both the high and learned-fidelity simulator, each simulation producing a binary outcome---$o_\mathrm{HF}$ and $o_\mathrm{LF}$---of either \textit{failure} or \textit{no failure}. The respective runtimes are denoted as $t_{\mathrm{HF}}$ and $t_{\mathrm{LF}}$. Because our objective is to identify failures in the system under test, a \textit{failure} outcome is considered a desirable result. Given the results from the two simulators, we can enumerate all four possible combinations of outcomes $o$: 1) \textit{true positive} (TP) when both simulations return \textit{failure}, 2) \textit{true negative} (TN) when both simulations return \textit{no failure}, 3) \textit{false positive} (FP) when the high-fidelity simulator returns \textit{no failure} and the learned-fidelity simulator returns \textit{failure}, and 4) \textit{false negative} (FN) when the high-fidelity simulator returns \textit{failure} and the learned-fidelity simulator returns \textit{no failure}. These outcome cases in combination with the runtimes are used for learning the distributions $p_\psi(\phi\mid s)$ and $p_\omega(\theta_{\mathrm{LF}}) $, a process which is further described in \cref{sec:fidelity_learning,sec:scenario_learning}.

\subsubsection{Evaluation}
The purpose of the evaluation is to assess the performance of the learned models after meta-training. In this phase, unlike during meta-training or testing, $\psi$ and $\omega$ are no longer updated. Furthermore, instead of sampling the fidelity settings, we greedily select them using the \textit{maximum a posteriori} estimate (MAP) of $p_\omega(\theta_{\mathrm{LF}})$, denoted as $\theta_{\mathrm{LF}}^\star$. Note that sampling $\phi$ from $p_\psi(\phi\mid s)$ is still required to avoid repeatedly simulating the same scenario instance. During the evaluation phase, we can assess how well the learned ${p_\psi(\phi\mid s)}$ and $p_\omega(\theta_{\mathrm{LF}})$ perform compared to a baseline, such as uniform sampling from the scenario settings and simulation only through the high-fidelity simulator. 

\subsubsection{Meta-Testing}
The purpose of the meta-testing phase is to evaluate whether the learned distribution over the fidelity settings $p_\omega(\theta_{\mathrm{LF}})$ generalizes well to a set of new scenarios $s_{n+1},\ldots,s_m$ and therefore provides a catalyst for faster learning of $p_\psi(\phi\mid s)~\text{for}~s\in\{s_{n+1},\ldots,s_m\}$. Note that the meta-testing phase is still a learning process, where $\psi$ and $\omega$ are updated, but $p_\omega(\theta_{\mathrm{LF}})$ is initialized with the posterior of $p_\omega(\theta_{\mathrm{LF}})$ from the meta-training phase while $p_\psi(\phi\mid s)$ is initialized with a uniform prior. The testing aspect focuses on whether the learned $p_\omega(\theta_{\mathrm{LF}})$ from meta-testing accelerates learning of $p_\psi(\phi\mid s)$ for new scenarios compared to initializing $p_\omega(\theta_{\mathrm{LF}})$ with a uniform distribution.

\subsection{Fidelity Setting Learning}\label{sec:fidelity_learning}
For learning the fidelity settings, an outcome is desirable when the high and learned-fidelity simulators produce consistent results (i.e., TP or TN). In addition, the ratio of learned-fidelity runtime over high-fidelity runtime should be less than a given compute budget $t_{\mathrm{LF}}/t_{\mathrm{HF}}\leq C_{\mathrm{budget}}$ with $C_{\mathrm{budget}}\in\left(0,1\right]$. By using the ratio $t_{\mathrm{LF}}/t_{\mathrm{HF}}$ rather than the raw runtime, we account for the fact that the simulated time between scenarios might vary. If $C_{\mathrm{budget}}$ is small, we require shorter runtimes of the learned-fidelity simulation compared to the high-fidelity simulation. 
\par We define a success for learning fidelity settings as 
% \begin{equation}
% \begin{aligned}
%     \mathcal{S}_{\mathrm{fidelity}} = \{o,t_\mathrm{LF},t_\mathrm{HF}\mid o\in\{\mathrm{TP},\mathrm{TN}\},&\\
%     t_\mathrm{LF}/  t_\mathrm{HF}\leq&C_\mathrm{budget}\}\label{eq:fidelity_success}
% \end{aligned}
% \end{equation}
\begin{equation}
    \mathcal{S}_{\mathrm{fidelity}} = \{\left(o,t_\mathrm{LF},t_\mathrm{HF}\right)\mid o\in\{\mathrm{TP},\mathrm{TN}\}, \frac{t_\mathrm{LF}}{t_\mathrm{HF}}\leq C_\mathrm{budget}\}\label{eq:fidelity_success}
\end{equation}
and a loss $\mathcal{L}_\mathrm{fidelity}$ consisting of $\left(0,t_\mathrm{LF},t_\mathrm{HF}\right)\not\in\mathcal{S}_\mathrm{fidelity}$.
A summary is provided in \cref{tab:failure_success}. If $C_{\mathrm{budget}}$ is chosen too small, we might encounter difficulty learning $p_\omega(\theta_{\mathrm{LF}})$ because the multi-armed bandit formulation requires successful trials as defined in \cref{eq:fidelity_success}. In most cases, there exists a lower bound $\underbar{C}_{\mathrm{budget}}$ where for any $C_{\mathrm{budget}}<\underbar{C}_\mathrm{bugdet}$ no successful trial can be encountered.
\par Under the assumption that each fidelity setting independently affects the behavior of a simulation, we model the problem as a multi-armed bandit problem \cite{robbins1952some}. Due to the independence assumptions between the different fidelity settings, we can solve $\lvert\theta_{\mathrm{LF}}\rvert$ independent multi-armed bandit problems, where the number of arms for each problem $i$ represents the number of values a fidelity setting $\theta_{LF,i}$ can take on. Although many fidelity settings are binary or categorical and well-suited for multi-armed bandit problems, other settings, such as distances, are continuous. By discretizing continuous fidelity settings, however, we can still use them with the multi-armed bandit framework.
\par The belief over the success for each possible fidelity setting value is represented by a beta distribution where $\alpha=n_{\mathrm{prior},\mathcal{S}}+n_{\mathcal{S}}$ and $\beta=n_{\mathrm{prior},\mathcal{L}}+n_{\mathcal{L}}$ with $n_{\mathcal{S}}$ and $n_{\mathcal{L}}$ being the counts of successes and losses as defined in \cref{tab:failure_success}, respectively. Prior knowledge about the distribution can be incorporated by adjusting $n_{\mathrm{prior},\mathcal{S}}$ and $n_{\mathrm{prior}_\mathcal{L}}$, whereas a uniform prior is equal to $n_{\mathrm{prior},\mathcal{S}}=n_{\mathrm{prior},\mathcal{L}}=1$.
\par To extract the optimal fidelity settings, the distribution over $\theta_{\mathrm{LF}}$ needs to be learned as accurately as possible. We use Thompson sampling \cite{thompson1933likelihood,agrawal2012analysis} to balance exploration and exploitation during training. For discrete fidelity settings, we can use Thompson sampling as described in the literature. For continuous fidelity settings, we use a stratified sampling scheme where the bin is sampled through Thompson sampling and the actual fidelity setting value is sampled within the bin according to either a uniform distribution (if uniform intervals were chosen) or a log-uniform distribution (if logarithmic intervals were chosen).
\par During the evaluation phase, there is no further need for exploration, and the \textit{optimal} fidelity settings can be selected using the greedy MAP estimate for each fidelity setting:
\begin{equation}
    \theta_{LF,i}^\star=\underset{j}{\arg\max}\frac{\alpha_j - 1}{
\alpha_j + \beta_j -2},
\end{equation}
where $i$ is indexing the fidelity setting and $j$ is indexing the possible values for each fidelity setting.
For continuous variables, we determine the expected value based on either a uniform or log-uniform distribution. 

\subsection{Learning Failure Scenarios}\label{sec:scenario_learning}
\begin{table}[!tb]
\caption{Success $\mathcal{S}$ and loss $\mathcal{L}$ for learning $p_\psi(\phi\mid s)$ and $p_\omega(\theta_{\mathrm{LF}})$}
\label{tab:failure_success}
\centering
\begin{tabular}{@{}lcr@{}}
    \toprule
        & \multicolumn{2}{@{}c}{Task}\\
    \cmidrule(l){2-3}
 &  Fidelity Settings & Scenario Parameters\\
    \midrule
    $\mathcal{S}$ & $\left( o\in\{\mathrm{TP, TN}\}\right)\land\left(t_{\mathrm{LF}}/ t_{\mathrm{HF}}\leq C_{\mathrm{budget}}\right)$ & $o\in\{\mathrm{TP},\mathrm{FN}\}$  \\
    $\mathcal{L}$  & $\left( o\in\{\mathrm{FP, FN}\}\right)\lor\left(t_{\mathrm{LF}}/ t_{\mathrm{HF}}> C_{\mathrm{budget}}\right)$ &$o\in\{\mathrm{TN},\mathrm{FP}\}$  \\
    \bottomrule
\end{tabular}
\end{table}
The method for learning failure scenarios---scenarios that are likely to lead to a failure of the system under test---is similar to the fidelity-learning framework described in \cref{sec:fidelity_learning} with three differences:
\begin{enumerate}
    \item A success $\mathcal{S}_{\mathrm{scenario}}$ is only dependent on the outcome and is defined when the outome $o$ is either TP or FN. Consequently a loss $\mathcal{L}_{\mathrm{scenario}}$ is defined when the outcome $o$ is either TN or FP.
    \item As each scenario type can have different parameters, a distribution $p_\psi(\phi)$ must be learned for each scenario $s$ which can be written as a conditional distribution $p_\psi(\phi\mid s)$.

    %As each scenario type can have different parameters, the distribution over $\phi$ is conditional on the scenario type $s$ while $p_\omega(\theta_{\mathrm{LF}})$ is scenario-agnostic.
    \item Instead of using the MAP estimate of the learned distribution during evaluation, we sample the scenarios from the learned distribution to prevent running the same scenario instance repeatedly.
\end{enumerate}

%Experiments
\section{\textsc{Experiments}}
We demonstrate the feasibility of our approach by using a state-of-the-art 3D autonomous driving simulator by Applied Intuition,\footnote{\url{https://www.appliedintuition.com}} a widely adopted platform in the autonomous driving industry. All experiments are run on a machine with an Intel Core i9-9900KF CPU, 64GB RAM, and a NVIDIA RTX 2080 Ti GPU. 
%Due to legal restrictions, we may not publish the code that interacts between the SAVME framework and the simulator, but 
SAVME is simulator-agnostic and we provide starter code with a generic simulator and instructions.\footnote{\url{https://github.com/sisl/SAVME.git}}
\subsection{System under Test}

\par We use a stack that consists of a localization sensor, a camera sensor, and a lidar sensor to demonstrate that our approach is capable of handling complex systems with many fidelity settings.  No motion planning is required because the desired lateral path is specified in the scenarios. For lateral path tracking, we use a Stanley controller \cite{hoffmann2007autonomous}, and for longitudinal control we use a PI controller that keeps the vehicle at a desired speed if no obstacle is detected. In the case of a predicted collision, a constant braking force is applied. The collision detection prediction is a two-stage process which begins with analyzing the camera image using a pre-trained YOLOv5s object detection network \cite{redmon2016you} trained on the widely-used COCO object detection dataset \cite{lin2014microsoft}.
\begin{figure}[!t]
\centering
    \begin{minipage}{0.45\linewidth}
        \resizebox{\linewidth}{!}{\begin{tikzpicture}

      %Set the boundaries
      \pgfmathsetmacro{\northboundary}{1.5}
      \pgfmathsetmacro{\westboundary}{0}
      \pgfmathsetmacro{\eastboundary}{7.5}
      \pgfmathsetmacro{\southboundary}{-5.5}
      
      %define the actual street
    \path[fill=black!30] (\westboundary,0) -- (3,0) arc (-90:0:1)  -- (4,\northboundary) -- (6,\northboundary) -- (6,1) arc (180:270:1) -- (\eastboundary,0) -- (\eastboundary,-2) -- (7,-2) arc (90:180:1) -- (6,\southboundary) -- (4,\southboundary) -- (4,-3) arc (0:90:1) -- (\westboundary,-2) -- cycle;

    %outer street markings
    \draw[color=white,very thick] (\westboundary,-0.1) -- (3,-0.1) arc (-90:0:1.1) -- (4.1,\northboundary);
    \draw[color=white,very thick] (5.9,\northboundary) -- (5.9,1) arc (180:270:1.1) -- (\eastboundary,-0.1);
    \draw[color=white,very thick] (\eastboundary,-1.9) -- (7,-1.9) arc (90:180:1.1) -- (5.9,\southboundary);
    \draw[color=white,very thick] (4.1,\southboundary) -- (4.1,-3) arc (0:90:1.1) -- (\westboundary,-1.9);

    %street markings
    \draw[color=trafficyellow, very thick] (\westboundary,-0.95) -- (3,-0.95);
    \draw[color=trafficyellow, very thick] (\westboundary,-1.05) -- (3,-1.05);

    \draw[color=trafficyellow, very thick] (\eastboundary,-0.95) -- (7,-0.95);
    \draw[color=trafficyellow, very thick] (\eastboundary,-1.05) -- (7,-1.05);

    \draw[color=trafficyellow, very thick] (4.95,\northboundary) -- (4.95,1);
    \draw[color=trafficyellow, very thick] (5.05,\northboundary) -- (5.05,1);

    \draw[color=trafficyellow, very thick] (4.95,\southboundary) -- (4.95,-3);
    \draw[color=trafficyellow, very thick] (5.05,\southboundary) -- (5.05,-3);

%intruder vehicle
\draw [color=coolgrey, very thick, dash pattern=on 8pt off 4pt] (2,-1.5) -- (3,-1.5) arc (-90:0:2.5) -- (5.5,\northboundary);
\node [sedan top,body color=coolgrey,window color=20black,minimum width=2cm] (ego) at (2,-1.5) {};

%ego vehicle
\draw [color=cardinalred, very thick, dash pattern=on 8pt off 4pt] (5.5,-4) -- (5.5,\northboundary);
\node [sedan top,body color=cardinalred,window color=20black,minimum width=2cm,rotate=90] at (5.5,-4) {};

  \end{tikzpicture}}
    \end{minipage}
\hfil
    \begin{minipage}{0.45\linewidth}
        \resizebox{\linewidth}{!}{\begin{tikzpicture}

      %Set the boundaries
      \pgfmathsetmacro{\northboundary}{1.5}
      \pgfmathsetmacro{\westboundary}{0}
      \pgfmathsetmacro{\eastboundary}{7.5}
      \pgfmathsetmacro{\southboundary}{-5.5}
      
      %define the actual street
    \path[fill=black!30] (\westboundary,0) -- (3,0) arc (-90:0:1)  -- (4,\northboundary) -- (6,\northboundary) -- (6,1) arc (180:270:1) -- (\eastboundary,0) -- (\eastboundary,-2) -- (7,-2) arc (90:180:1) -- (6,\southboundary) -- (4,\southboundary) -- (4,-3) arc (0:90:1) -- (\westboundary,-2) -- cycle;

    %outer street markings
    \draw[color=white,very thick] (\westboundary,-0.1) -- (3,-0.1) arc (-90:0:1.1) -- (4.1,\northboundary);
    \draw[color=white,very thick] (5.9,\northboundary) -- (5.9,1) arc (180:270:1.1) -- (\eastboundary,-0.1);
    \draw[color=white,very thick] (\eastboundary,-1.9) -- (7,-1.9) arc (90:180:1.1) -- (5.9,\southboundary);
    \draw[color=white,very thick] (4.1,\southboundary) -- (4.1,-3) arc (0:90:1.1) -- (\westboundary,-1.9);

    %street markings
    \draw[color=trafficyellow, very thick] (\westboundary,-0.95) -- (3,-0.95);
    \draw[color=trafficyellow, very thick] (\westboundary,-1.05) -- (3,-1.05);

    \draw[color=trafficyellow, very thick] (\eastboundary,-0.95) -- (7,-0.95);
    \draw[color=trafficyellow, very thick] (\eastboundary,-1.05) -- (7,-1.05);

    \draw[color=trafficyellow, very thick] (4.95,\northboundary) -- (4.95,1);
    \draw[color=trafficyellow, very thick] (5.05,\northboundary) -- (5.05,1);

    \draw[color=trafficyellow, very thick] (4.95,\southboundary) -- (4.95,-3);
    \draw[color=trafficyellow, very thick] (5.05,\southboundary) -- (5.05,-3);

    %ego vehicle
    \draw [color=cardinalred, very thick, dash pattern=on 8pt off 4pt] (2,-1.5) -- (3,-1.5) arc (-90:0:2.5) -- (5.5,\northboundary);
    \node [sedan top,body color=cardinalred,window color=20black,minimum width=2cm] (ego) at (2,-1.5) {};

    %intruder vehicle
    \draw [color=coolgrey, very thick, dash pattern=on 8pt off 4pt] (5.5,-4) -- (5.5,\northboundary);
    \node [sedan top,body color=coolgrey,window color=20black,minimum width=2cm,rotate= 90] at (5.5,-4) {};

  \end{tikzpicture}}
    \end{minipage}
\vskip 0.1cm
    \begin{minipage}{0.45\linewidth}
    \centering
        {\footnotesize (1)}
    \end{minipage}
\hfil
    \begin{minipage}{0.45\linewidth}
    \centering
        {\footnotesize (2)}
    \end{minipage}
\vskip 0.40cm
    \begin{minipage}{0.45\linewidth}
        \resizebox{\linewidth}{!}{\begin{tikzpicture}

      %Set the boundaries
      \pgfmathsetmacro{\northboundary}{4}
      \pgfmathsetmacro{\westboundary}{0}
      \pgfmathsetmacro{\eastboundary}{7.5}
      \pgfmathsetmacro{\southboundary}{-3.5}
      
      %define the actual street
    \path[fill=black!30] (\westboundary,0) -- (3,0) arc (-90:0:1)  -- (4,\northboundary) -- (6,\northboundary) -- (6,1) arc (180:270:1) -- (\eastboundary,0) -- (\eastboundary,-2) -- (7,-2) arc (90:180:1) -- (6,\southboundary) -- (4,\southboundary) -- (4,-3) arc (0:90:1) -- (\westboundary,-2) -- cycle;

    %outer street markings
    \draw[color=white,very thick] (\westboundary,-0.1) -- (3,-0.1) arc (-90:0:1.1) -- (4.1,\northboundary);
    \draw[color=white,very thick] (5.9,\northboundary) -- (5.9,1) arc (180:270:1.1) -- (\eastboundary,-0.1);
    \draw[color=white,very thick] (\eastboundary,-1.9) -- (7,-1.9) arc (90:180:1.1) -- (5.9,\southboundary);
    \draw[color=white,very thick] (4.1,\southboundary) -- (4.1,-3) arc (0:90:1.1) -- (\westboundary,-1.9);

    %street markings
    \draw[color=trafficyellow, very thick] (\westboundary,-0.95) -- (3,-0.95);
    \draw[color=trafficyellow, very thick] (\westboundary,-1.05) -- (3,-1.05);

    \draw[color=trafficyellow, very thick] (\eastboundary,-0.95) -- (7,-0.95);
    \draw[color=trafficyellow, very thick] (\eastboundary,-1.05) -- (7,-1.05);

    \draw[color=trafficyellow, very thick] (4.95,\northboundary) -- (4.95,1);
    \draw[color=trafficyellow, very thick] (5.05,\northboundary) -- (5.05,1);

    \draw[color=trafficyellow, very thick] (4.95,\southboundary) -- (4.95,-3);
    \draw[color=trafficyellow, very thick] (5.05,\southboundary) -- (5.05,-3);

     %intruder vehicle
    \draw [color=coolgrey, very thick, dash pattern=on 8pt off 4pt] (2,-1.5) -- (3,-1.5) arc (-90:0:2.5) -- (5.5,\northboundary);
    \node [sedan top,body color=coolgrey,window color=20black,minimum width=2cm] (ego) at (2,-1.5) {};

    %ego vehicle
    \draw [color=cardinalred, very thick, dash pattern=on 8pt off 4pt] (4.5,2.5) -- (4.5,\southboundary);
    \node [sedan top,body color=cardinalred,window color=20black,minimum width=2cm,rotate=-90] at (4.5,2.5) {};

  \end{tikzpicture}}
    \end{minipage}
\hfil
    \begin{minipage}{0.45\linewidth}
        \resizebox{\linewidth}{!}{\begin{tikzpicture}

      %Set the boundaries
      \pgfmathsetmacro{\northboundary}{4}
      \pgfmathsetmacro{\westboundary}{0}
      \pgfmathsetmacro{\eastboundary}{7.5}
      \pgfmathsetmacro{\southboundary}{-3.5}
      
      %define the actual street
    \path[fill=black!30] (\westboundary,0) -- (3,0) arc (-90:0:1)  -- (4,\northboundary) -- (6,\northboundary) -- (6,1) arc (180:270:1) -- (\eastboundary,0) -- (\eastboundary,-2) -- (7,-2) arc (90:180:1) -- (6,\southboundary) -- (4,\southboundary) -- (4,-3) arc (0:90:1) -- (\westboundary,-2) -- cycle;

    %outer street markings
    \draw[color=white,very thick] (\westboundary,-0.1) -- (3,-0.1) arc (-90:0:1.1) -- (4.1,\northboundary);
    \draw[color=white,very thick] (5.9,\northboundary) -- (5.9,1) arc (180:270:1.1) -- (\eastboundary,-0.1);
    \draw[color=white,very thick] (\eastboundary,-1.9) -- (7,-1.9) arc (90:180:1.1) -- (5.9,\southboundary);
    \draw[color=white,very thick] (4.1,\southboundary) -- (4.1,-3) arc (0:90:1.1) -- (\westboundary,-1.9);

    %street markings
    \draw[color=trafficyellow, very thick] (\westboundary,-0.95) -- (3,-0.95);
    \draw[color=trafficyellow, very thick] (\westboundary,-1.05) -- (3,-1.05);

    \draw[color=trafficyellow, very thick] (\eastboundary,-0.95) -- (7,-0.95);
    \draw[color=trafficyellow, very thick] (\eastboundary,-1.05) -- (7,-1.05);

    \draw[color=trafficyellow, very thick] (4.95,\northboundary) -- (4.95,1);
    \draw[color=trafficyellow, very thick] (5.05,\northboundary) -- (5.05,1);

    \draw[color=trafficyellow, very thick] (4.95,\southboundary) -- (4.95,-3);
    \draw[color=trafficyellow, very thick] (5.05,\southboundary) -- (5.05,-3);

    %ego vehicle
    \draw [color=cardinalred, very thick, dash pattern=on 8pt off 4pt] (2,-1.5) -- (3,-1.5) arc (-90:0:2.5) -- (5.5,\northboundary);
    \node [sedan top,body color=cardinalred,window color=20black,minimum width=2cm] (ego) at (2,-1.5) {};

    %intruder vehicle
    \draw [color=coolgrey, very thick, dash pattern=on 8pt off 4pt] (4.5,2.5) -- (4.5,\southboundary);
    \node [sedan top,body color=coolgrey,window color=20black,minimum width=2cm,rotate=-90] at (4.5,2.5) {};

  \end{tikzpicture}}
    \end{minipage}
\vskip 0.1cm
    \begin{minipage}{0.45\linewidth}
    \centering
        {\footnotesize (3)}
    \end{minipage}
\hfil
    \begin{minipage}{0.45\linewidth}
    \centering
        {\footnotesize (4)}
    \end{minipage}
\vskip 0.40cm
    \begin{minipage}{0.45\linewidth}
        \resizebox{\linewidth}{!}{\begin{tikzpicture}
         \pgfmathsetmacro{\westboundary}{0}
         \pgfmathsetmacro{\eastboundary}{10}

         %define the pavement
         \path[fill=black!30] (\westboundary,1) -- (\eastboundary,1) -- (\eastboundary,-1) -- (\westboundary,-1) -- cycle;

         %street boundaries
         \draw[color=trafficyellow,very thick] (\westboundary,0.9) -- (\eastboundary,0.9);
         \draw[color=white,very thick] (\westboundary,-0.9) -- (\eastboundary,-0.9);

         %lane separation
         \draw[color=white, very thick, dash pattern=on 10pt off 10pt ] (\westboundary,0) -- (\eastboundary,0);

         %ego
         \draw [color=cardinalred, very thick, dash pattern=on 8pt off 4pt] (1.3,-0.5) -- (8,-0.5);
         \node [sedan top,body color=cardinalred,window color=20black,minimum width=2cm] (ego) at (1.3,-0.5) {};

         %intruder
         \draw [color=coolgrey, very thick, dash pattern=on 8pt off 4pt] (2.5,0.5) -- (4,0.5) to[out=0,in=180] (9,-0.5) -- (\eastboundary,-0.5);  % arc (90:70:3) arc (250:270:3) ;
         \node [sedan top,body color=coolgrey,window color=20black,minimum width=2cm,rotate=0] at (2.5,0.5) {};

     \end{tikzpicture}}
    \end{minipage}
\hfil
    \begin{minipage}{0.45\linewidth}
        \resizebox{\linewidth}{!}{\begin{tikzpicture}
         \pgfmathsetmacro{\westboundary}{0}
         \pgfmathsetmacro{\eastboundary}{10}

         %define the pavement
         \path[fill=black!30] (\westboundary,1) -- (\eastboundary,1) -- (\eastboundary,-1) -- (\westboundary,-1) -- cycle;

         %street boundaries
         \draw[color=trafficyellow,very thick] (\westboundary,0.9) -- (\eastboundary,0.9);
         \draw[color=white,very thick] (\westboundary,-0.9) -- (\eastboundary,-0.9);

         %lane separation
         \draw[color=white, very thick, dash pattern=on 10pt off 10pt ] (\westboundary,0) -- (\eastboundary,0);

%intruder
\draw [color=coolgrey, very thick, dash pattern=on 8pt off 4pt] (1.3,-0.5) -- (8,-0.5);
\node [sedan top,body color=coolgrey,window color=20black,minimum width=2cm] (ego) at (1.3,-0.5) {};

%ego
\draw [color=cardinalred, very thick, dash pattern=on 8pt off 4pt] (2.5,0.5) -- (4,0.5) to[out=0,in=180] (9,-0.5) -- (\eastboundary,-0.5);  arc (90:70:3) arc (250:270:3);
\node [sedan top,body color=cardinalred,window color=20black,minimum width=2cm,rotate=0] at (2.5,0.5) {};

     \end{tikzpicture}}
    \end{minipage}
\vskip 0.1cm
    \begin{minipage}{0.45\linewidth}
    \centering
        {\footnotesize (5)}
    \end{minipage}
\hfil
    \begin{minipage}{0.45\linewidth}
    \centering
        {\footnotesize (6)}
    \end{minipage}
\vskip 0.40cm
    \begin{minipage}{0.45\linewidth}
        \resizebox{\linewidth}{!}{     \begin{tikzpicture}
         \pgfmathsetmacro{\westboundary}{0}
         \pgfmathsetmacro{\eastboundary}{10}

         %define the pavement
         \path[fill=black!30] (\westboundary,1) -- (\eastboundary,1) -- (\eastboundary,-1) -- (\westboundary,-1) -- cycle;

         %street boundaries
         \draw[color=trafficyellow,very thick] (\westboundary,0.9) -- (\eastboundary,0.9);
         \draw[color=white,very thick] (\westboundary,-0.9) -- (\eastboundary,-0.9);

         %lane separation
         \draw[color=white, very thick, dash pattern=on 10pt off 10pt ] (\westboundary,0) -- (\eastboundary,0);

%ego
\draw [color=cardinalred, very thick, dash pattern=on 8pt off 4pt] (1.3,-0.5) -- (5,-0.5);
\node [sedan top,body color=cardinalred,window color=20black,minimum width=2cm] (ego) at (1.3,-0.5) {};

%intruder
\draw [color=coolgrey, very thick, dash pattern=on 8pt off 4pt] (7,-0.5) -- (8,-0.5); to[out=0,in=180] (9,-0.5) -- (\eastboundary,-0.5); 
\node [sedan top,body color=coolgrey,window color=20black,minimum width=2cm,rotate=0] at (6,-0.5) {};
         
     \end{tikzpicture}}
    \end{minipage}
\hfil
    \begin{minipage}{0.45\linewidth}
        \resizebox{\linewidth}{!}{\begin{tikzpicture}
    \pgfmathsetmacro{\westboundary}{0}
     \pgfmathsetmacro{\eastboundary}{10}
     \pgfmathsetmacro{\southboundary}{-2.5}
     % \pgfmathsetmacro{\southboundary}{6}
     \def\nolines{9}

     %draw pavement
     \path[fill=black!30] (\westboundary,1.3) -- (\eastboundary,1.3) -- (\eastboundary,\southboundary) -- (\westboundary,\southboundary) -- cycle;

     %lane markers
    \draw[color=trafficyellow, very thick, dash pattern=on 10pt off 10pt ] (\westboundary,1) -- (\eastboundary,1);
     \draw[color=white,very thick] (\westboundary,0) -- (\eastboundary,0);

     %\draw parking lot
     \draw[color=white,very thick] ($(\westboundary,\southboundary + 0.3)$) -- ($(\eastboundary,\southboundary + 0.3)$);
     \foreach \i in {1,...,\nolines}
     {
        \draw[color=white, very thick] ($(\i,\southboundary+0.3)$) -- (\i,-0.3);
     }

     %ego
     \draw [color=cardinalred, very thick, dash pattern=on 8pt off 4pt] (1.3,0.5) -- (5.5,0.5);
     \node [sedan top,body color=cardinalred,window color=20black,minimum width=2cm] (ego) at (1.3,0.5) {};

     %intruder
     \draw [color=coolgrey, very thick, dash pattern=on 8pt off 4pt] (7.5,-0.8) arc (10:90:1.58) -- (\eastboundary,0.5);
     \node [sedan top,body color=coolgrey,window color=20black,minimum width=2cm,rotate=-80] at (7.5,-0.8) {}; 
     
    \end{tikzpicture}}
    \end{minipage}
\vskip 0.1cm
    \begin{minipage}{0.45\linewidth}
    \centering
        {\footnotesize (7)}
    \end{minipage}
\hfil
    \begin{minipage}{0.45\linewidth}
    \centering
        {\footnotesize (8)}
    \end{minipage}
\vskip 0.40cm
    \begin{minipage}{0.45\linewidth}
        \resizebox{\linewidth}{!}{\begin{tikzpicture}
         \pgfmathsetmacro{\westboundary}{0}
         \pgfmathsetmacro{\eastboundary}{10}

         %define the pavement
         \path[fill=black!30] (\westboundary,1) -- (\eastboundary,1) -- (\eastboundary,-1) -- (\westboundary,-1) -- cycle;

         %street boundaries
         \draw[color=white,very thick] (\westboundary,0.9) -- (\eastboundary,0.9);
         \draw[color=white,very thick] (\westboundary,-0.9) -- (\eastboundary,-0.9);

         %lane separation
         \draw[color=trafficyellow, very thick ] (\westboundary,0.05) -- (\eastboundary,0.05);
         \draw[color=trafficyellow, very thick] (\westboundary,-0.05) -- (\eastboundary,-0.05);

%ego
\draw [color=cardinalred, very thick, dash pattern=on 8pt off 4pt] (1.3,-0.5) -- (5,-0.5);
\node [sedan top,body color=cardinalred,window color=20black,minimum width=2cm] (ego) at (1.3,-0.5) {};

%intruder
\draw [color=coolgrey, very thick, dash pattern=on 8pt off 4pt] (5.5,-0.5) -- (7,-0.5); to[out=0,in=180] (9,-0.5) -- (\eastboundary,-0.5); 
\node [sedan top,body color=coolgrey,window color=20black,minimum width=2cm,rotate=0] at (7.5,-0.5) {};
         
     \end{tikzpicture}}
    \end{minipage}
\hfil
    \begin{minipage}{0.45\linewidth}
        \resizebox{\linewidth}{!}{\begin{tikzpicture}
   %define the pavement
  \path[fill=black!30] (0,-1) -- (0,0) -- (10,0) -- (10,-1) to[in=0, out=180] (6,-1.8) -- (5,-1.8) arc (90:140:5) -- +(140:1) arc (140:115:6) to[out=25,in=0] (2.3,-1) -- cycle;

  %street markings
  \draw[color=trafficyellow,very thick] (0,-0.1) -- (10,-0.1);
  \draw[color=white, very thick] (10,-0.9) to[in=0, out=180] (6,-1.7) -- (5,-1.7) arc (90:140:5.1);
  \draw[color=white, very thick] (0,-0.9) -- (5,-0.9);
  \draw[color=white, very thick, dash pattern=on 10pt off 10pt ] (5,-0.9) -- (9.5,-0.9);
  \draw[white, very thick] (5,-0.9) arc (90:113:5.9) coordinate (endwhitearc);
  \draw[color=trafficyellow,very thick] (endwhitearc) arc (113.5:140:5.9);

  %ego
  \draw [color=cardinalred, very thick, dash pattern=on 8pt off 4pt] (10,-0.5) to[in=0,out=180] (6,-1.3) -- (5,-1.3) arc (90:130:5.5);
  \node [sedan top,body color=cardinalred,window color=20black,minimum width=2cm,rotate=39] (ego) at (1.5,-2.58) {};

  %intruder
  \draw [color=coolgrey, very thick, dash pattern=on 8pt off 4pt] (2,-0.5) -- (9,-0.5);
  \node [sedan top,body color=coolgrey,window color=20black,minimum width=2cm,rotate=0] at (2,-0.5) {};

\end{tikzpicture}}
    \end{minipage}
\vskip 0.1cm
    \begin{minipage}{0.45\linewidth}
    \centering
        {\footnotesize (9)}
    \end{minipage}
\hfil
    \begin{minipage}{0.45\linewidth}
    \centering
        {\footnotesize (10)}
    \end{minipage}

\caption{Pre-crash scenarios. Scenarios 1 through 8 are used for meta-training, while scenarios 9 and 10 are used for meta-testing. The red vehicle represents the ego vehicle while the gray vehicle is the obstacle.}\label{fig:pre_crash_scenarios}

\end{figure}
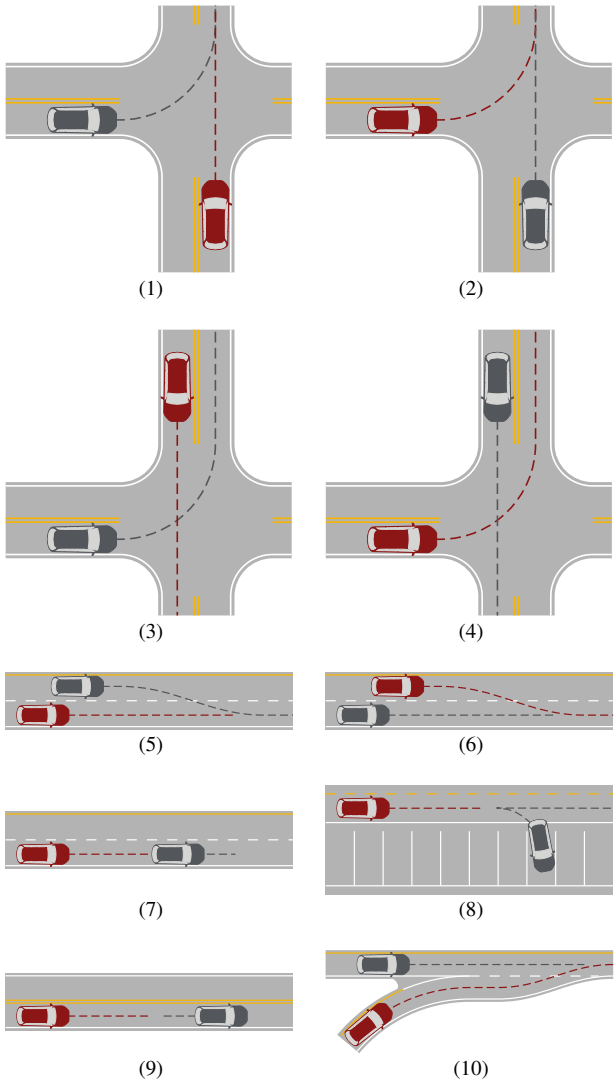
\par If an object of the category \textit{vehicle} or \textit{person} is detected, the lidar signal is filtered according to the discovered bounding box and the weighted centroid of the filtered lidar pointcloud is taken as measurement for the obstacle's position. The weights are proportional to the intensity of the point as reported by the lidar sensor. We predict the closest distance between the ego vehicle and the obstacle based on a first-order point-mass model which is fitted to the current and previous obstacle position. \comment{make footnote that we outsource this to the repo} If the predicted minimal distance is less than a threshold of one car length, the brake event is triggered. We choose this safety buffer to account for noisy measurements and the inaccuracies of the first-order model. A more detailed description can be found in the repository.
\begin{table}[!b]
\centering
\caption{Fidelity settings with $\theta_\mathrm{HF}$ highlighted}\label{tab:fidelity_settings}
\begin{tabular}{@{}lllr@{}}
\toprule
Sensor & Fidelity Setting & Type & Values/Range \\ \midrule
- & simulation rate& discrete & \{2, 4, 6, 8, \textbf{10}\} Hz\\
camera & bloom level & discrete &  \{\textbf{high}, low\}\\
camera & disable bloom & discrete &  \{true, \textbf{false}\}\\
camera & disable lighting& discrete &  \{true, \textbf{false}\}\\
camera & disable shadows & discrete & \{true, \textbf{false}\}\\
camera & disable lens model & discrete & \{true, \textbf{false}\}\\
camera & disable depth of field & discrete & \{true, \textbf{false}\}\\
camera & disable shot noise & discrete & \{true, \textbf{false}\}\\
camera & view distance & continuous & [10, \textbf{5000}] m\\
camera & near clipping distance & continuous & [\textbf{0.2}, 20] m\\
lidar & disable shot noise & discrete & \{true, \textbf{false}\}\\
lidar & disable ambient effects & discrete & \{true, \textbf{false}\}\\
lidar & disable translucency & discrete & \{true, \textbf{false}\}\\
lidar & subsample count & discrete & \{1, 2, 3, 4, \textbf{5}\}\\
lidar & raytracing bounces & discrete & \{0, 1, 2, 3, \textbf{4}\}\\
lidar & near clipping distance & continuous & [\textbf{0.2}, 20] m\\ \bottomrule
\end{tabular}
\end{table}
\subsection{Scenarios}
\par Instead of the NHTSA pre-crash typology \cite{najm2007pre} which is based on crashes between human-operated cars, we use a pre-crash typology from crash reports of incidents involving autonomous vehicles \cite{liu2021crash}. A total of 10 scenario types where the autonomous vehicle plays an active role in the accident are used for our assessment. No scenarios in which the autonomous vehicle is passive are included, such as instances where it is rear-ended. Out of the 10 scenarios, we use 8 scenarios for the meta-training phase and 2 scenarios for the meta-testing phase. The scenarios are depicted in \cref{fig:pre_crash_scenarios} where scenarios 1 through 8 are the scenarios that are used for the meta-training while scenarios 9 and 10 are used for meta-testing.

\subsection{Fidelity Settings.}
To demonstrate the feasibility of our framework in a high-dimensional, mixed categorical and continuous fidelity space, we use the 16 fidelity settings in \cref{tab:fidelity_settings} alongside the definition for $\theta_\mathrm{HF}$.

\subsection{Baseline and Experiment Goals}\label{sec:baseline_experiment_goals}
As a baseline, we sample the scenario settings $\phi$ from a uniform distribution and evaluate those using the high-fidelity settings $\theta_{\mathrm{HF}}$ that correspond to the simulator's recommended settings. We can formulate two experimental goals based on the results from the meta-training and meta-testing phases:
\begin{enumerate}
    % \item Using the same scenarios $s_1,\ldots,s_n$ and the learned distributions $p(\phi\mid s_1),\ldots,p(\phi\mid s_n)$, but only use the low-fidelity simulator with $\theta_{\mathrm{LF}}^\star$, will significantly speed up the training and lead to the discovery of more failures. We can determine an approximate break-even point when we found more failures despite the parallel training with a high and low-fidelity simulator during the meta-training phase.
    \item The evaluation of the meta-training phase reveals how well the proposed framework can detect scenarios that lead to failures, i.e., learning ${p_\psi(\phi\mid s)}$, while also making each simulation run faster by adjusting the the fidelity settings, i.e., learning $p_\omega(\theta_{\mathrm{LF}})$. At the end of the meta-training phase, it is possible to calculate the speedup factor, which denotes the increase in the number of detected failures within the same time span as compared to the baseline. Of further interest is the break-even point, or time at which we detect the same number of failures using our dual high and learned-fidelity simulators than we did with the baseline. 
    %---despite the computationally expensive training---we find more failures than using the baseline that does not involve the dual-simulator setup.
    % \item Looking at a new set of scenarios $s_{n+1},\ldots,s_m$, we have no information about $p(\phi\mid s_{n+1}),\ldots,p(\phi\mid s_m)$, but given the learned optimal fidelity settings $\theta_{\mathrm{LF}}^\star$, it is possible to check whether $p(\phi\mid s_{n+1}),\ldots,p(\phi\mid s_m)$ can be learned faster using only the low-fidelity simulator during meta-testing.
    \item While $p_\psi(\phi\mid s)$ is conditional on the scenario, $p_\omega(\theta_{\mathrm{LF}})$ is scenario-agnostic and therefore used across all scenarios during the meta-training phase. During the meta-testing phase, we evaluate whether using the learned $p_\omega(\theta_{\mathrm{LF}})$ from meta-training as a prior helps to speed up learning $p_\psi(\phi\mid s)$ on unseen scenarios.
\end{enumerate}
Our findings for both goals are presented in \cref{sec:results}.

%Results
\section{\textsc{Results}}\label{sec:results}

% \begin{figure}[!ht]
%     \centering
%     \includegraphics[width=\linewidth]{figures/meta_test.pdf}
%     \caption{Found failures over compute time during meta-testing phase.}
%     \label{fig:meta_testing}
% \end{figure}

\subsection{Meta-Training}
By isolating the results of the meta-training phase, we can understand the suitability of the multi-armed bandit approach for simultaneously learning $p_\psi(\phi\mid s)$ and $p_\omega(\theta_{\mathrm{LF}})$. We train both $p_\psi(\phi\mid s)$ and $p_\omega(\theta_{\mathrm{LF}})$ using scenarios 1 through 8 as shown in \cref{fig:pre_crash_scenarios} for 500 iterations where the probability $p(s)$ is uniform. We evaluate the learned $p_\psi(\phi\mid s)$ and $p_\omega(\theta_{\mathrm{LF}})$ using 100 evaluations with $\phi\sim p_\psi(\phi\mid s)$ and $\theta_{\mathrm{LF}}^\star=\underset{\theta_{\mathrm{LF}}}{\arg\max}~p_\omega(\theta_{\mathrm{LF}})$. For each $C_{\mathrm{budget}}\in\{0.2,0.3,0.4\}$, we calculate the TP-rate (how many failures we would've found only using the learned-fidelity simulator), the mean relative runtime of the learned-fidelity simulator when compared against the high-fidelity simulator, and the relative speedup from the baseline. All results are shown in \cref{tab:meta_training_results}. We first note that for all $C_{\mathrm{budget}}$, the mean learned-fidelity cost is remarkably similar given the different bounds. 
%Despite different values for $C_{\mathrm{budget}}$, the fidelity settings that lead to the highest fidelity success rate where a success is defined in \cref{eq:fidelity_success} lead to a similar mean cost.
% We interpret this as an efficient way for maximizing the probability of a successful fidelity that passes the criterion in \cref{eq:fidelity_success}. 
Second, we note the significant performance drop for $C_{\mathrm{budget}}=0.2$ which indicates that 0.2 approaches $\underbar{C}_{\mathrm{budget}}$. Finally, an almost 18 times speedup relative baseline is achieved with $C_{\mathrm{budget}}=0.3$. In other words, at evaluation, using the learned-fidelity simulator, we found almost 18 times as many failures within the same runtime.
\begin{table}[!b]
\sisetup{detect-weight=true}
\centering
\caption{Results from evaluation after meta-training}
\label{tab:meta_training_results}
\begin{tabular}{@{}lS[table-format=3.2,table-number-alignment = center]S[table-format=3.2,table-number-alignment = center]S[table-format=2.2,table-number-alignment = right]@{}}
% \begin{tabular}{llll}
\toprule
{$C_{\mathrm{budget}}$} & {TP-Rate} & {Mean LF Cost}& {Speedup} \\ \midrule
{baseline}        & 0.17                  & 1.00         & 1.00    \\
0.4          & 0.41                  & 0.26         & 9.32   \\
0.3          & \bfseries 0.74                 & 0.25         & \bfseries 17.99   \\
0.2          & 0.29                  & \bfseries 0.22         & 7.86   \\ \bottomrule
\end{tabular}
\end{table}
\begin{figure}[!b]
    \centering
    \includegraphics[width=\linewidth]{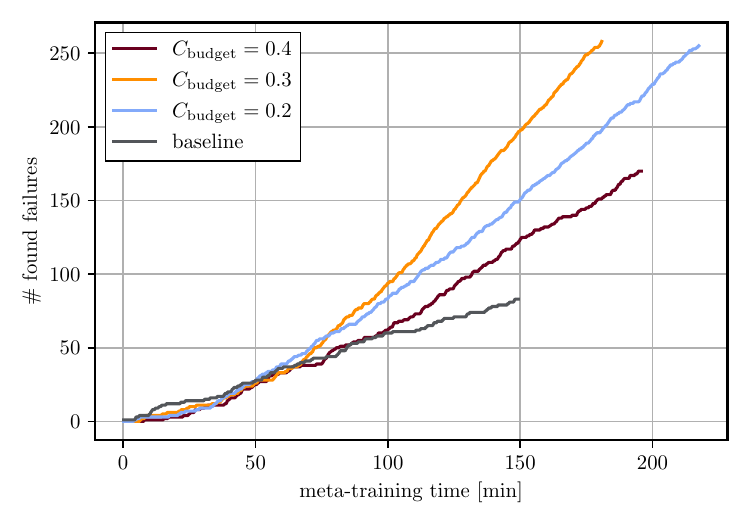}
    \caption{Found failures over runtime during the meta-training phase. Each experiment is 500 iterations long. For all $C_{\mathrm{budget}}$, the break-even point is reached before the end of the training.}
    \label{fig:breakeven_plot}
\end{figure}
\par Concurrently executing the high-fidelity and learned-fidelity simulators during meta-training incurs significant computational cost. Thus, it is important to determine the runtime threshold at which our approach exhibits a higher incidence of failures compared to the baseline, also called the break-even point. The break-even point can either occur during or after training. \Cref{fig:breakeven_plot} shows the failures over the runtime for the baseline and for the meta-training phase with $C_{\mathrm{budget}}=\{0.2,0.3,0.4\}$, each executed for 500 iterations. We conclude that the break-even point occurs before the end of the meta-training phase between \SI[group-minimum-digits=4]{50}{min} and \SI[group-minimum-digits=4]{80}{min} for all $C_{\mathrm{budget}}$. In other words, despite the concurrent usage of the high and learned-fidelity simulators, we find more failures during the meta-training phase than the baseline.

\par
\begin{figure}[!b]
  \centering
  \begin{subfigure}{0.48\linewidth}
    \centering
    \includegraphics[width=\linewidth]{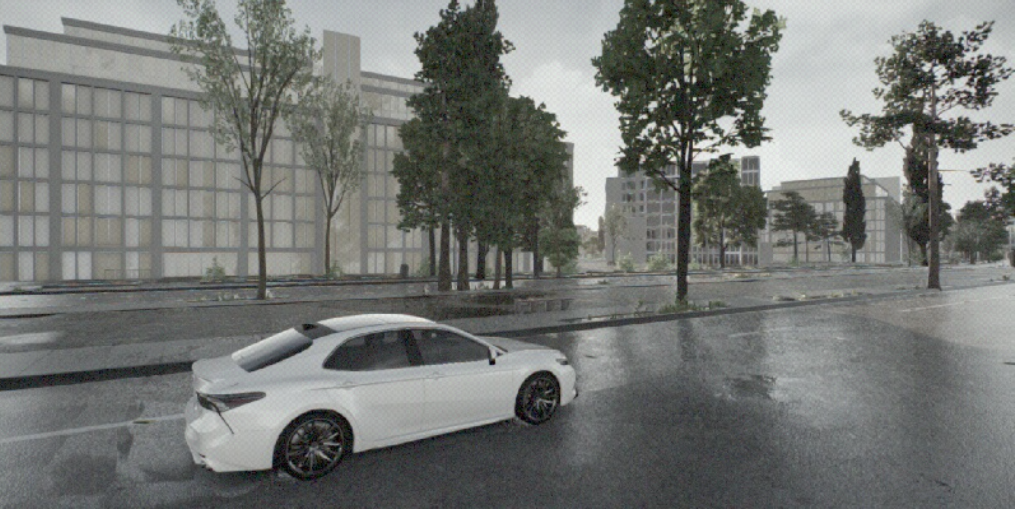}
    \caption{Camera image with $\theta_\mathrm{HF}^{\phantom{\star}}$}
    \label{fig:subfig1}
  \end{subfigure}
  \hfill
  \begin{subfigure}{0.48\linewidth}
    \centering
    \includegraphics[width=\linewidth]{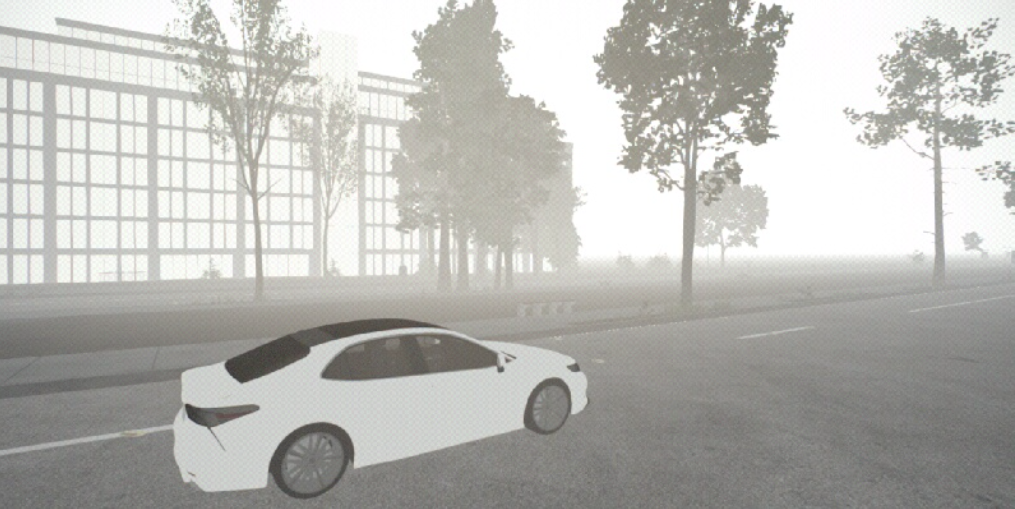}
    \caption{Camera image with $\theta_\mathrm{LF}^{\star}$}
    \label{fig:subfig2}
  \end{subfigure}
  \caption{Comparison of images from the camera sensor between $\theta_\mathrm{HF}^{\phantom{\star}}$ and $\theta_\mathrm{LF}^{\star}$ for $C_\mathrm{budget}=0.3$. The differences can be seen especially well in the maximum view distance and reflections.}
  \label{fig:camera_hf_lf_comparison}
\end{figure}

 \Cref{tab:MAP_fidelity_settings} shows the learned fidelity settings. While some settings differ based on $C_{\mathrm{budget}}$, most settings remain the same. Finally, \cref{fig:camera_hf_lf_comparison} shows an example of a camera image contrasting the difference between the high and learned-fidelity settings.
\subsection{Meta-Testing}
\begin{table}[!b]
\centering
\caption{Learned fidelity settings}\label{tab:MAP_fidelity_settings}
\begin{adjustbox}{max width=\linewidth}
\begin{tabular}{@{}llllr@{}}
\toprule
Sensor & Fidelity Setting & $C_{b.}=0.4$ & $C_{b.}=0.3$ & $C_{b.}=0.2$ \\ \midrule
- & simulation rate& 2 Hz & 2 Hz & 2 Hz\\
camera & bloom level & low & high & low\\
camera & disable bloom & true & true & false\\
camera & disable lighting& true & false & false\\
camera & disable shadows &true & true & true\\
camera & disable lens model &false & false & true\\
camera & disable depth of field & true & false & false\\
camera & disable shot noise & true & true & false\\
camera & view distance & 65.70 m& 285.97 m& 65.70 m\\
camera & near clipping distance & 3.67 m& 7.22 m & 0.58 m\\
lidar & disable shot noise & false & true & false\\
lidar & disable ambient effects & false & false & false\\
lidar & disable translucency &true & false & false\\
lidar & subsample count & 3 & 2 & 1\\
lidar & raytracing bounces & 0 & 0 & 1\\
lidar & near clipping distance & 1.68 m & 1.68 m & 13.57 m\\ 
\bottomrule

\end{tabular}
\end{adjustbox}
\end{table}

\begin{figure}[!tb]
    \centering
    \includegraphics[width=\linewidth]{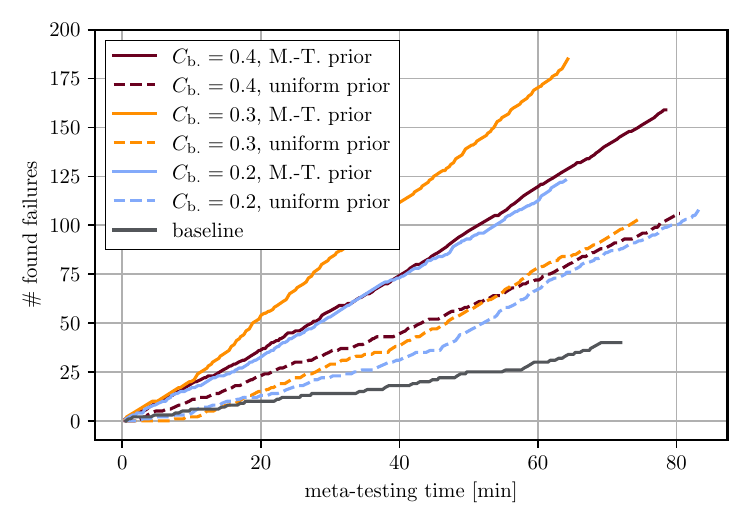}
    \caption{Found failures over runtime during meta-testing. Each experiment is 200 iterations long. Using the posterior of $p_\omega(\theta_{\mathrm{LF}})$ from meta-training as prior during meta-testing resulted in faster learning of $p_\psi(\phi\mid s)$ for all $C_\mathrm{budget}$.}
    \label{fig:meta_testing}
\end{figure}

For meta-testing, we use scenarios 9 and 10 from \cref{fig:pre_crash_scenarios} to evaluate the effect of using the posterior distribution $p_\omega(\theta_{\mathrm{LF}})$ from the meta-training phase as prior for the meta-testing phase. We compare against the case of using a uniform prior for $p_\omega(\theta_{\mathrm{LF}})$ as well as the previously introduced baseline. The meta-testing phase has a duration of 200 iterations where we record the runtime and the found failures. The results are depicted in \cref{fig:meta_testing}. Using the learned prior for $p_\omega(\theta_{\mathrm{LF}})$ results in more found failures for all compute budgets $C_{\mathrm{budget}}=\{0.2,0.3,0.4\}$ during the meta-testing phase when compared to using a uniform prior. We thus arrive at the conclusion that utilizing the posterior $p_\omega(\theta_{\mathrm{LF}})$ obtained during the meta-training phase as the prior for $p_\omega(\theta_{\mathrm{LF}})$ while learning $p_\psi(\phi\mid s)$ for new scenarios $s_{n+1},\ldots,s_m$ exhibits a beneficial impact on the rate of learning. For our experiments, we observe an increase of 50 to 100\%. We also included the baseline that was introduced in \cref{sec:baseline_experiment_goals} to demonstrate the two levels from this study: The comparison of the uniform prior with the baseline is the setup during meta-training while the comparison between a uniform prior and the learned prior is the setup for meta-testing. \Cref{fig:meta_testing} can be seen as the summary of this entire study. We demonstrate that even with a uniform prior on $p_\omega(\theta_\mathrm{LF})$, the SAVME framework performs better than the baseline, but using the learned prior on $p_\omega(\theta_\mathrm{LF})$, leads to an even more increased speedup.

\subsection{Limitations}
We acknowledge two primary limitations inherent in our present approach. First, because we are using a multi-armed bandit framework, we are restricted to discrete or discretized scenario parameters and fidelity settings. This limitation could be overcome by using a more general Dirichlet process formulation. Second, assuming independence within and between the scenario parameters and fidelity settings might be an oversimplification that can necessitate the use of more advanced techniques involving probabilistic graphical models. While it is true that many applications may not be significantly impacted by these limitations, we believe they serve as a valuable foundation for expanding the potential applications of the SAVME framework in the future.

\section{\textsc{Conclusion}}
This paper presents an efficient, falsification-based validation approach for autonomous systems using meta-learning in conjunction with a Bayesian multi-armed bandit formulation. Our framework is unique as we are approaching efficient safety validation from the falsification perspective as well as from the efficient simulation perspective. In addition, with our approach, the learned scenario-agnostic fidelity settings can be used to accelerate the falsification process for new scenarios, which represents the meta aspect of our method. The SAVME framework's source code is open-source and requires minimal effort to apply to other problems.
\par In our experiments we use a state-of-the-art 3D driving simulator and a driving stack with camera and lidar sensors.  The scenarios come from an AV-specific pre-crash typology while the simulation setup provides 16 fidelity settings. During the evaluation following meta-training, our framework demonstrates on average an almost 18-fold reduction in the time required to detect a failure. Furthermore, we find that despite the concurrent use of two simulators, more failures can be discovered even during the training phase when compared to the baseline that only uses one simulator. During the meta-learning phase, we observe that using the fidelity settings obtained during meta-training as a prior, increases the falsification process up to a factor of two.

\addtolength{\textheight}{-12cm}   % This command serves to balance the column lengths

%%%%%%%%%%%%%%%%%%%%%%%%%%%%%%%%%%%%%%%%%%%%%%%%%%%%%%%%%%%%%%%%%%%%%%%%%%%%%%%%

%%%%%%%%%%%%%%%%%%%%%%%%%%%%%%%%%%%%%%%%%%%%%%%%%%%%%%%%%%%%%%%%%%%%%%%%%%%%%%%%

%%%%%%%%%%%%%%%%%%%%%%%%%%%%%%%%%%%%%%%%%%%%%%%%%%%%%%%%%%%%%%%%%%%%%%%%%%%%%%%%
% \input{sections/appendix}

\section*{\textsc{Acknowledgements}}

The authors would like to express their sincere gratitude to Allstate for their generous funding support through the Stanford Center for AI Safety as well as to Applied Intuition for the software access and technical support.

%%%%%%%%%%%%%%%%%%%%%%%%%%%%%%%%%%%%%%%%%%%%%%%%%%%%%%%%%%%%%%%%%%%%%%%%%%%%%%%%

% References are important to the reader; therefore, each citation must be complete and correct. If at all possible, references should be commonly available publications.

\bibliographystyle{IEEEtran}
\bibliography{references}

\end{document}